\documentclass[11pt]{article}

\usepackage[preprint]{acl}

\usepackage{times}
\usepackage{latexsym}

\usepackage{amsmath, amsfonts, amssymb}
\usepackage[T1]{fontenc}

\usepackage[utf8]{inputenc}

\usepackage{microtype}

\usepackage{inconsolata}

\usepackage{graphicx}
\usepackage{listings}
\usepackage{multirow}
\usepackage{booktabs}
\usepackage{makecell}
\usepackage{subcaption}
\usepackage{enumitem}
\usepackage{tcolorbox}
\tcbuselibrary{breakable}
\usepackage{dsfont}
\usepackage[table]{xcolor}
\usepackage{fvextra}

\title{InfoDensity: Rewarding Information-Dense Traces for Efficient Reasoning}

\author{
Chengwei Wei\textsuperscript{$\diamondsuit$}, 
Jung-jae Kim \textsuperscript{$\diamondsuit$}, 
Longyin Zhang \textsuperscript{$\diamondsuit$}, 
Shengkai Chen \textsuperscript{$\diamondsuit$}, 
Nancy F. Chen\textsuperscript{$\diamondsuit,\dag$}
\\
\textsuperscript{$\diamondsuit$}Institute for Infocomm Research (I$^2$R), A*STAR, Singapore\\
\textsuperscript{$\dag$}Centre for Frontier AI Research (CFAR), A*STAR, Singapore\\
\texttt{wei\_chengwei@a-star.edu.sg} \\
}

\begin{document}
\maketitle
\begin{abstract}
Large Language Models (LLMs) with extended reasoning capabilities often generate verbose and redundant reasoning traces, incurring unnecessary computational cost. While existing reinforcement learning approaches address this by optimizing final response length, they neglect the quality of intermediate reasoning steps, leaving models vulnerable to reward hacking. 
We argue that verbosity is not merely a length problem, but a symptom of poor intermediate reasoning quality. To investigate this, we conduct an empirical study tracking the per-token predictive entropy of large reasoning models across reasoning trajectories. We find that high-quality reasoning traces exhibit two consistent properties: \textit{low uncertainty convergence} and \textit{fast uncertainty descent}. 
These findings suggest that high-quality reasoning traces are informationally dense, that is, reasoning steps contribute to reaching a low uncertainty level relative to the total reasoning length.
Motivated by this, we propose \textbf{InfoDensity}, a reward framework for RL training that captures both properties through a single suffix-max envelope of the entropy trajectory, weighted by a length scaling term that favors achieving equivalent quality more concisely. 
Experiments on mathematical and general reasoning benchmarks demonstrate that InfoDensity outperforms state-of-the-art baselines on the accuracy--efficiency trade-off.
\end{abstract}

\section{Introduction}
Recent reasoning-focused Large Language Models (LLMs), often termed Large Reasoning Models (LRMs) \cite{guo2025deepseek, yang2025qwen3}, have demonstrated remarkable capabilities across diverse domains by generating multi-step Chain-of-Thought (CoT) traces~\cite{wei2022chain}. Despite their effectiveness, LRMs tend to produce verbose and redundant reasoning traces, leading to unnecessary token generation and elevated computational cost~\cite{chen2024not, zhu2025towards, luo2025o1}. 

To address this, reinforcement learning (RL) with length-based rewards has emerged as a prominent approach for encouraging concise reasoning. One line of work incorporates length penalties directly into the RL objective to incentivize shorter reasoning traces~\cite{team2025kimi, arora2025training}, while another line of work determines explicit length targets during training, either through progressively tightened token budgets~\cite{aggarwal2025l1, hou2025thinkprune} or by estimating problem difficulty to set adaptive length thresholds~\cite{liu2025learn, shen2025dast}.

However, these approaches share a key limitation: by optimizing only the length and final answer correctness of the reasoning trace, they leave the quality of intermediate reasoning steps unsupervised. We argue that verbosity is not merely a length problem, but a symptom of a deeper issue: incorrect or redundant intermediate steps reflect poor reasoning quality. Without supervising reasoning quality, models are vulnerable to reward hacking, where they learn to produce superficially concise responses that satisfy length constraints without generating correct and complete reasoning traces~\cite{yeo2025demystifying, shrivastava2025sample}. 

Measuring the quality of reasoning traces is a non-trivial challenge. Process Reward Models (PRMs)~\cite{lightman2024let} introduced step-level supervision for evaluating reasoning processes, and subsequent work has automated such signals via Monte Carlo Tree Search~\cite{luo2024improve} or multi-dimensional rubrics~\cite{do2025defines}. 
More recently, information-theoretic frameworks derive the signal directly from the model's own probability estimates.
Rather than relying on labeled step-level data, they quantify reasoning quality either by tracking the model's own per-token predictive entropy during generation~\citep{zhu2025uncertainty} or by measuring how much each step reduces the model's uncertainty about the final answer~\citep{ton2025understanding, guo2025measuring}. Both enable the identification of low-quality reasoning purely from the model's own probability estimates. Inspired by this, we ask whether such information-theoretic signals can be incorporated as reward signals in RL training to guide models toward efficient and high-quality reasoning.

In this work, we first conduct an empirical study on how information-theoretic signals characterize the quality of reasoning traces using the model's own per-token predictive entropy. Specifically, we use reasoning trace correctness as a proxy for quality, treating traces that arrive at correct answers via coherent intermediate reasoning as high-quality, and those that fail to do so as low-quality. We find that correct traces exhibit two consistent properties: they converge to lower entropy by the end of the reasoning phase (\textbf{low uncertainty convergence}) and reduce entropy more sharply across the trajectory (\textbf{fast uncertainty descent}). By contrast, incorrect traces sustain higher entropy throughout the reasoning phase. Together, these findings suggest that high-quality reasoning traces are informationally dense: 
reasoning steps contribute to reaching a low uncertainty level relative to the total reasoning length.

Motivated by these findings, we propose \textbf{InfoDensity}, a reward framework for RL training that directly measures reasoning quality by encouraging high information density throughout the reasoning trace. Unlike prior entropy-based rewards that capture only a single aspect of the entropy trajectory, InfoDensity uses a single suffix-max aggregator over the per-token entropy trajectory to jointly capture both low uncertainty convergence and fast uncertainty descent as a unified measure of reasoning quality, further weighted by a length scaling term that favors more concise reasoning at equal quality. Experiments on mathematical and general reasoning benchmarks show that InfoDensity matches or outperforms state-of-the-art baselines in accuracy while significantly reducing token usage, achieving a strong trade-off between accuracy and efficiency.

Our main contributions are as follows:
\begin{itemize}[itemsep=1pt, topsep=1pt]
\item We conduct an empirical study of entropy trajectories across correct and incorrect reasoning traces, identifying two trajectory-level properties, low uncertainty convergence and fast uncertainty descent, that distinguish high-quality from low-quality reasoning traces.
\item We propose InfoDensity, an entropy trajectory-based reward framework that supervises both the quality and conciseness of reasoning traces. InfoDensity achieves a better balance between accuracy and generation length than state-of-the-art baselines, with consistent results across diverse reasoning benchmarks and models.
\item We open-source the InfoDensity model, training, and evaluation pipelines to enable reproducibility and support further research in efficient reasoning.\footnote{Code and models will be available at \url{https://github.com/amao0o0/InfoDensity}.}
\end{itemize}

\section{Related Work}
\paragraph{Efficient Reasoning.} 
A growing body of work targets the computational overhead of LRM chain-of-thought~\citep{fengefficient, zhu2025towards} via prompt-driven methods~\citep{xu2025chain, aytes2025sketch, ma2025reasoning}, length-efficient supervised fine-tuning~\citep{xia2025tokenskip, kang2025c3ot}, and decoding-stage early termination~\citep{zhang2025consistent, yong2026think, yang2025dynamic}. Meanwhile, RL with length-based rewards has become particularly prominent, spanning direct or accuracy-conditional length penalties~\citep{team2025kimi, arora2025training, zhang2025grpo, luo2025o1}, token-budget schedules~\citep{aggarwal2025l1, hou2025thinkprune}, problem-adaptive length targets~\citep{shen2025dast, xiang2025just, li2025aalc, yi2026shorterbetter}, and rewards augmented by signals from an auxiliary scoring model~\citep{deng2025rea}.

More recently, several methods use the model's own predictive entropy as an additional reward signal. PEAR~\citep{huang2025pear} averages per-token entropy separately within the thinking and answer phases and rewards a low thinking-phase mean. Step Entropy~\citep{li2025compressing} uses per-step entropy sums offline to prune low-information steps. Concurrent with this work, ETR~\citep{xiong2026etr} rewards a momentum-aggregated descent trend of stepwise entropy. InfoDensity instead uses a single suffix-max reward over the entropy trajectory that jointly captures both low uncertainty convergence and fast uncertainty descent, rewarding information-dense reasoning.

\paragraph{Evaluation of Reasoning Traces.}
Evaluating reasoning quality in LLMs has evolved beyond final-answer accuracy to assess intermediate steps. Process Reward Models (PRMs)~\citep{lightman2024let} rely on large-scale human annotations, and later work automates this supervision via Monte Carlo Tree Search~\citep{luo2024improve} or rubric-based scoring~\citep{do2025defines}. Information-theoretic approaches instead derive the signal directly from the model's own probability estimates, either as step-wise information gain via teacher-forced conditional entropy~\citep{ton2025understanding, guo2025measuring} or as aggregated per-token entropy~\citep{zhu2025uncertainty}. Our work goes beyond evaluation by identifying two trajectory-level properties of entropy: low uncertainty convergence and fast uncertainty descent, and directly incorporating them as reward signals in RL training.

\section{Information-theoretic Analysis of Reasoning Traces} \label{sec:analysis}

\subsection{Preliminary} \label{sec:analysis_prelim}

We formalize the per-token entropy, per-step entropy, and per-step uncertainty reduction computation for a reasoning trace produced by an LRM. Let $X$ be an input problem and $Y = (Y_1, Y_2, \ldots, Y_T)$ be the reasoning trace, where $Y_t$ is the $t$-th intermediate reasoning step containing $K_t$ tokens $Y_t = (y_{t,1}, \ldots, y_{t,K_t})$. The model with parameters $\theta$ generates these tokens autoregressively from the conditional distribution $p_\theta(\cdot \mid X, Y_{<t})$.

\paragraph{Per-token entropy.} Let $c_{t,k} = (X, Y_{<t}, y_{t,<k})$ denote the context preceding the $k$-th token of step $Y_t$. The token-level entropy of the model's predictive distribution over the vocabulary $\mathcal{V}$ is
\[
H(y_{t,k} \mid c_{t,k})
\;=\; -\!\sum_{v \in \mathcal{V}}
p_\theta(v \mid c_{t,k})\, \log p_\theta(v \mid c_{t,k}).
\]

\paragraph{Per-step entropy.}
The step-level entropy $H_t$ is the mean of per-token entropies across the $K_t$ tokens of step $Y_t$:
\[
H_t \;=\; \frac{1}{K_t} \sum_{k=1}^{K_t} H(y_{t,k} \mid c_{t,k}),
\]
representing the average per-token uncertainty the LRM exhibits while generating step $Y_t$. 

\paragraph{Per-step uncertainty reduction.}
Beyond the absolute entropy level, a natural measure of step quality is the step-by-step uncertainty reduction
\[
\Delta H_t \;=\; H_{t-1} - H_t, \qquad t = 2, \ldots, T,
\]
which captures how much the model's uncertainty decreased during step $Y_t$. A positive $\Delta H_t$ indicates that the step committed the model to a more decisive predictive distribution, providing information that enables the model to 
make progress without redundant deliberation.
The implementation for computing per-token and per-step entropy is provided in Appendix~\ref{sec:appendix_entropy_impl}.

\subsection{Step-level and Trajectory-level Analysis}
\label{sec:ig_eval}

We investigate whether the entropy and uncertainty reduction computed from an LRM can serve as a practical signal for reasoning quality. A reasoning step can be low-quality in two ways: it may be incorrect, introducing errors that derail the reasoning process, or redundant, contributing little new information per generated token. Both manifest as a failure to reduce the model's uncertainty. We therefore use step correctness as a proxy for reasoning quality and examine whether $H_t$ and $\Delta H_t$ can distinguish correct from incorrect steps. We first examine these signals at the step level, then shift to the entropy trajectory of a full reasoning trace.

\paragraph{Dataset.} We construct a step-quality dataset from ProcessBench~\citep{zheng2025processbench}, which provides human-annotated step-level correctness labels across diverse math problems of varying difficulty. We select samples whose corresponding problems exist in the original source datasets (e.g., GSM8K, MATH) and have a ground-truth answer. 
For step-level analysis, we identify the first incorrect step in each incorrect reasoning trace and exclude all subsequent steps to ensure each step's entropy and uncertainty reduction are measured without interference from preceding errors. For trajectory-level analysis, we exclude the final steps of all reasoning traces because those steps typically consist of an answer-stating conclusion (e.g., ``Therefore, the answer is '') whose tokens reflect formatting rather than reasoning content. Dataset statistics are summarized in Table~\ref{tab:dataset_correctness}.

\begin{table}[b]                                                           
\centering                                                                 
\small                                                                     
\caption{Statistics of the ProcessBench subset used for step-level and trajectory-level analysis.}      
\resizebox{\linewidth}{!}{
\begin{tabular}{lccc}                                                   
\toprule                                                                   
\textbf{Dataset} & \textbf{\#Questions} & \textbf{\%Correct} & \textbf{Avg.\ Steps} \\ 
\midrule                                                                   
GSM8K         & 400 & 48.3 & 3.9 \\
MATH          & 117 & 43.6 & 4.7 \\                 
OlympiadBench & 1000 & 33.9 & 5.8 \\               
Omni-MATH     & 999 & 24.1 & 4.8 \\                                      
\bottomrule                                                                
\end{tabular}}                                                 
\label{tab:dataset_correctness}                                            
\end{table} 

\begin{figure}[tbh]
    \centering
    \includegraphics[width=0.99\linewidth]{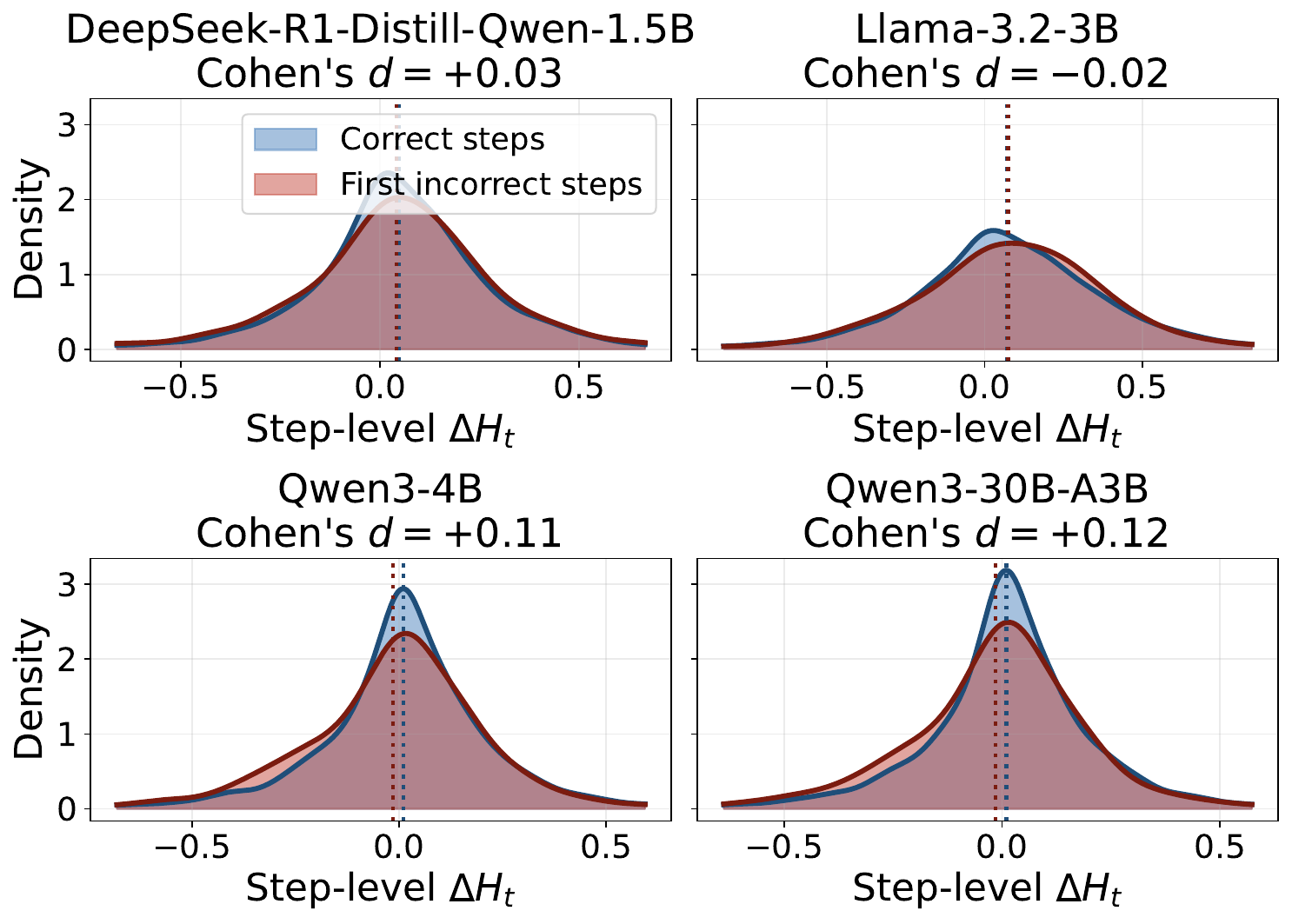}
    \caption{Step-level uncertainty reduction ($\Delta H_t$) distributions for correct (blue) and first-incorrect (red) steps in the ProcessBench subset across four models. Dotted vertical lines mark the per-class means.}
    \label{fig:ig_dist}
\end{figure}

\paragraph{Challenges of step-level uncertainty reduction.}
For each reasoning step $t$, a positive $\Delta H_t$ should indicate that the step reduced the model's per-token uncertainty, while a low or negative value might signal an unproductive or failed step. We test this across five base models spanning four families and 1.5B--30B scale: DeepSeek-R1-Distill-Qwen-1.5B~\citep{guo2025deepseek}, Llama-3.2-3B~\citep{grattafiori2024llama}, Gemma-3-4B-pt~\citep{kamath2025gemma}, Qwen3-4B and Qwen3-30B-A3B~\citep{yang2025qwen3}.

Across all five models and four ProcessBench subsets (GSM8K, MATH, OlympiadBench, Omni-MATH), step-level $\Delta H_t$ is empirically too noisy to serve as a standalone classifier of step correctness. We quantify the overlap between the $\Delta H_t$ distributions of correct and first-incorrect steps using Cohen's $d$, which measures the standardized distance between the two means. As shown for four representative models in Figure~\ref{fig:ig_dist}, the distributions are heavily overlapping, with paired Cohen's $d$ lying in $[-0.02, +0.12]$, well below the conventional 0.2 threshold for a small effect. Treating $\Delta H_t$ as a step-correctness classifier (with higher $\Delta H_t$ predicting correct steps), we find that it yields ROC curves clustering tightly around the chance diagonal, with AUC in $[0.49, 0.53]$ (Figure~\ref{fig:ig_roc} in Appendix~\ref{sec:appendix_step_level}). The low classification performance indicates that $\Delta H_t$ at a single step is too noisy to serve as a robust error detector.

\begin{figure*}[tbh]
    \centering
    \includegraphics[width=0.95\linewidth]{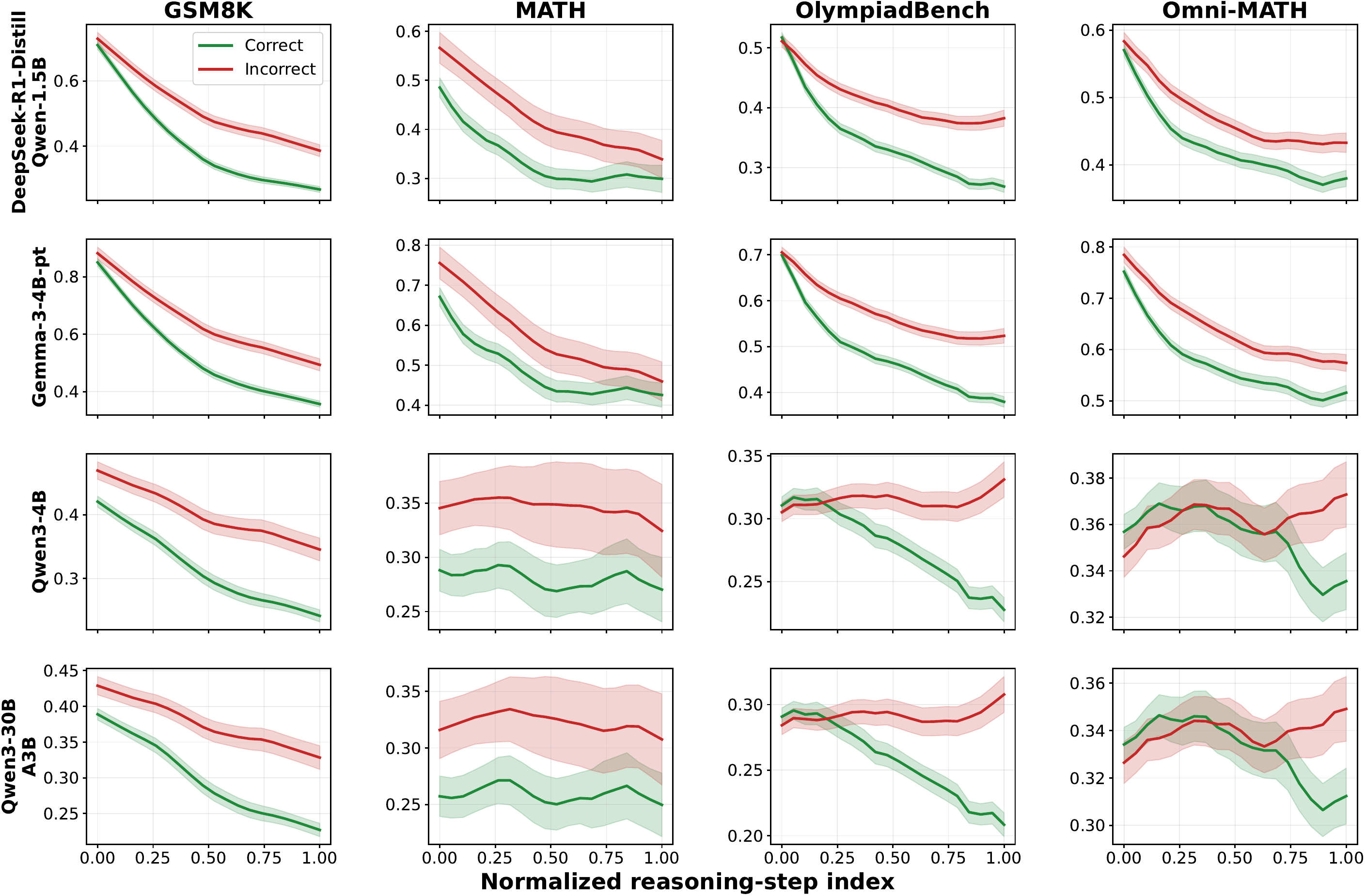}
    \caption{Step-level entropy trajectory across four datasets: GSM8K, MATH, OlympiadBench, and Omni-MATH. Each curve shows the mean entropy over normalized reasoning steps, with shaded regions indicating $\pm 1$ standard error.}
    \label{fig:entropy_trajectory}
\end{figure*}

\paragraph{From local steps to full trajectories.}
We therefore shift focus to the shape of the full entropy trajectory $\{H_1, H_2, \ldots, H_T\}$. Figure~\ref{fig:entropy_trajectory} presents entropy trajectories for four representative models on four ProcessBench subsets (full five-model panel in Appendix~\ref{sec:appendix_step_level}). Since reasoning traces vary in length, we normalize all entropy trajectories to a fixed length via linear interpolation before averaging. Green and red lines correspond to correct and incorrect reasoning traces, respectively.

A clear divergence emerges between correct and incorrect traces within each (model, dataset) cell. Across all five models and four datasets, correct traces consistently exhibit a decline in step-level entropy and converge to lower values than incorrect traces by the end of the reasoning phase, though the decline is weaker for the two Qwen3 base models on MATH and Omni-MATH. We discuss this phenomenon in detail in Appendix~\ref{sec:appendix_qwen3_quirk}. 

The shape of the incorrect trajectories differs across families: on R1-Distill-Qwen-1.5B, Llama-3.2-3B, and Gemma-3-4B-pt, the incorrect traces also descend. The two Qwen-family base models instead show incorrect trajectories that remain flat or slightly ascend on the two harder datasets (OlympiadBench, Omni-MATH), suggesting that on these harder problems, Qwen-family models start with low-entropy first steps and then gradually accumulate uncertainty as the flawed reasoning trace progresses.

\paragraph{Characterizing high-quality reasoning.} The entropy trajectories reveal two complementary properties that separate correct from incorrect reasoning traces:

\begin{itemize}[itemsep=2pt, topsep=2pt]
\item \textbf{Low uncertainty convergence}: Correct traces converge to substantially lower per-token entropy by the end of the reasoning phase, while incorrect traces end at substantially higher entropy.
\item \textbf{Fast uncertainty descent}: Correct traces reduce per-token entropy more sharply across the trajectory, while incorrect traces decline slowly or remain high.
\end{itemize}

Together, these two properties imply that high-quality reasoning traces are \emph{information-dense}: across the trajectory, uncertainty is reduced more and settles into a low-entropy level by the conclusion, whereas flawed traces dissipate uncertainty more slowly or not at all. These properties jointly motivate the trajectory-level reward we propose in Section~\ref{sec:infodensity}, where a single aggregator is constructed to capture both the convergence and the descent simultaneously.

\section{InfoDensity} \label{sec:infodensity}

Building on our trajectory-level observations, we propose \textbf{InfoDensity}, a reward framework designed to promote efficient reasoning. The name reflects its core idea: it captures information quality relative to reasoning length, combining a quality score that measures how completely a trace resolves uncertainty
with a length scaling term that favors achieving this more concisely.

\paragraph{Information Quality.}
The two trajectory-level properties identified in Section~\ref{sec:ig_eval}, low uncertainty convergence and fast uncertainty descent, motivate two complementary operations on the step-level entropy trajectory $\{H_1, H_2, \ldots, H_{T}\}$.

Low uncertainty convergence requires the trajectory's tail entropy to be low. We capture this with the right-to-left running maximum of the trajectory, which we call the \emph{suffix-max envelope}, a curve that bounds the trajectory from above by taking, at each position, the maximum entropy from that position to the end:
\begin{multline}\label{eq:suffix}
\mathrm{suffix}_t = \max(H_t,\, H_{t+1},\, \ldots,\, H_{T}), \\
t = 1, 2, \ldots, T.
\end{multline}
A high entropy value near the trajectory's end raises $\mathrm{suffix}_t$ at every earlier position through the $\max$
operation. Conversely, a trajectory that stays low at the tail keeps the envelope low throughout.

Fast uncertainty descent is then captured by taking the length-normalized mean of the envelope. For a monotonically descending
trajectory, $\mathrm{suffix}_t = H_t$ at every position (each future maximum coincides with the current value), so the envelope mean reduces to the trajectory mean $\bar H = (1/T)\sum_t H_t$. A faster descent leaves more steps at low entropy, yielding a smaller $\bar H$ and thus a smaller envelope mean.

The two operations together define the quality reward:
\begin{equation}\label{eq:rquality}
R_{\mathrm{quality}}(\tau) = 1 - \frac{1}{T \log K}\sum_{t=1}^{T}
\mathrm{suffix}_t,
\end{equation}
where $\log K$ (we set $K = 20$) is the maximum entropy of a uniform distribution over $K$ tokens, used as a fixed normalization anchor. A high $R_{\mathrm{quality}}$ therefore requires both a low tail and a fast descent.

\paragraph{Length Scaling.}
For the length scaling term, we adopt a group-relative formulation that requires no predefined length target. Following GRPO-LEAD~\citep{zhang2025grpo}, we define:

\begin{equation}\label{eq:length}
R_{L}(\tau) \;=\; \exp\!\left(-\lambda \cdot \frac{L(\tau) - \mu_L}{\sigma_L}\right)
\end{equation}

where $L(\tau)$ is the response length, $\mu_L, \sigma_L$ are the mean and standard deviation of lengths within the rollout group, and $\lambda > 0$ is a hyperparameter controlling the strength of length scaling. Shorter-than-average responses yield $R_L > 1$, while longer-than-average ones yield $R_L < 1$.

\paragraph{The InfoDensity Formulation.}
The final reward is the product of the quality score and the length scaling term:
\begin{equation}
R_{\mathrm{InfoDensity}}(\tau) \;=\; R_{\mathrm{quality}}(\tau) \cdot R_{L}(\tau).
\end{equation}
$R_{\mathrm{InfoDensity}}$ is applied only to traces with correct final answers. The final reward for a trace $\tau$ is:
\begin{equation}
R(\tau) \;=\; \begin{cases}
R_{\mathrm{InfoDensity}}(\tau) & \text{if } \tau \text{ is correct,} \\
0 & \text{otherwise.}
\end{cases}
\end{equation}

\section{Results}
\subsection{Experiment Setup}

\paragraph{Base Models.}
We evaluate InfoDensity on widely used LRMs spanning 1.5B--8B parameters, drawn from two model families: the DeepSeek-R1-Distill-Qwen series (DSR-1.5B and DSR-7B)~\citep{guo2025deepseek} and the Qwen3-8B~\citep{yang2025qwen3}. The original model performance is also evaluated as a reference to quantify the improvement in both accuracy and token efficiency.

\paragraph{Baseline Methods.}
We compare InfoDensity against six representative baselines:
\textbf{1) DEER}~\citep{yang2025dynamic} is a training-free method that triggers confidence-based early exit during reasoning. \textbf{2) NoThink}~\citep{ma2025reasoning} is a training-free prompting strategy that bypasses the thinking phase to force an immediate answer. \textbf{3) LCPO}~\citep{aggarwal2025l1} is a length-constrained RL baseline that trains the model to match a prompt-specified token budget. \textbf{4) O1-Pruner}~\citep{luo2025o1} applies length-harmonizing RL fine-tuning that rewards length reduction relative to a reference trajectory. \textbf{5) PEAR}~\citep{huang2025pear} uses phase-level entropy as the reward signal, penalizing high entropy during thinking and encouraging low entropy in the answer. \textbf{6) ETR}~\citep{xiong2026etr} is the closest concurrent work, rewarding a momentum-aggregated trend of stepwise entropy descent along the reasoning trace.

\paragraph{Training.}
All baseline methods and InfoDensity are trained on the 7000-problem subset of DeepMath-103K~\citep{he2025deepmath} (difficulty levels 5--10) released by~\citet{xiong2026etr}. We train InfoDensity with a rollout group size of $n{=}5$, maximum response length of 16384 tokens, learning rate $1{\times}10^{-5}$, batch size 32, and KL coefficient $\beta{=}0.005$. The InfoDensity length-scaling coefficient is set to $\lambda{=}0.05$ unless otherwise specified. 

\paragraph{Evaluation Setup.}
To evaluate reasoning ability across both mathematical and general reasoning QA domains, we report results on four benchmarks: \textbf{AMC23}~\citep{maa_amc23}, \textbf{AIME24}~\citep{aime24}, and \textbf{MATH500}~\citep{hendrycks2021measuring} for competition-level mathematical reasoning, and \textbf{GPQA-Diamond} (GPQA-D)~\citep{rein2023gpqa}, a graduate-level expert QA benchmark covering physics, chemistry, and biology, for general scientific reasoning beyond mathematics. This benchmark suite is widely used in recent reasoning work~\citep{guo2025deepseek, xiong2026etr}, enabling direct comparison with concurrent and prior baselines under an identical evaluation protocol.

\begin{table*}[h]
\centering
\caption{Evaluation results on four benchmarks across three models. $\uparrow$/$\downarrow$ indicate higher/lower-is-better. The best AES per dataset and overall within each model block is \textbf{bolded}. The second best is \underline{underlined}.}
\resizebox{1\linewidth}{!}
{\begin{tabular}{l | c c c | c c c | c c c | c c c || c c c}
\toprule
\multirow{3}{*}{\textbf{Method}} & \multicolumn{3}{c|}{\textbf{AMC23}} & \multicolumn{3}{c|}{\textbf{AIME24}} &
\multicolumn{3}{c|}{\textbf{MATH500}} & \multicolumn{3}{c||}{\textbf{GPQA-D}} & \multicolumn{3}{c}{\textbf{Overall}}
\\
\cmidrule(lr){2-4} \cmidrule(lr){5-7} \cmidrule(lr){8-10} \cmidrule(lr){11-13} \cmidrule(lr){14-16} & Acc$\uparrow$
& Len$\downarrow$ & AES$\uparrow$ & Acc$\uparrow$ & Len$\downarrow$ & AES$\uparrow$& Acc$\uparrow$ & Len$\downarrow$
& AES$\uparrow$ & Acc$\uparrow$ & Len$\downarrow$ & AES$\uparrow$ & Acc$\uparrow$ & Len$\downarrow$ & AES$\uparrow$
\\
\midrule
\rowcolor{gray!20}
\multicolumn{16}{c}{\textit{DeepSeek-R1-Distill-Qwen-1.5B}} \\
\midrule
Original & 50.0 & 9.6k & 0.00 & 20.0 & 14.3k & 0.00 & 69.6 & 6.2k & 0.00 & 6.6 & 14.4k & 0.00 & 36.5 & 11.1k & 0.00 \\
DEER & 57.5 & 7.0k & 0.72 & 23.3 & 11.8k & 0.67 & 65.6 & 3.1k & 0.21 & 5.6 & 12.5k & -0.63 & 38.0 & 8.6k & 0.35 \\
NoThink & 62.5 & 2.4k & 1.49 & 6.7 & 7.6k & -2.86 & 71.8 & 1.3k & \underline{0.89} & 9.1 & 1.5k & 2.03 & 37.5 & 3.2k & 0.79 \\
LCPO & 67.5 & 7.6k & 1.25 & 30.0 & 12.3k & 1.64 & 74.2 & 5.2k & 0.36 & 8.6 & 13.5k & 0.97 & 45.1 & 9.7k & 0.83 \\
O1-Pruner & 60.0 & 8.4k & 0.72 & 16.7 & 13.9k & -0.81 & 68.2 & 6.3k & -0.12 & 8.1 & 14.2k & 0.71 & 38.2 & 10.7k & 0.17 \\
PEAR & 60.0 & 6.5k & 0.92 & 36.7 & 11.7k & \textbf{2.68} & 77.0 & 4.4k & 0.61 & 11.1 & 12.8k & \underline{2.16} & 46.2 & 8.8k & 1.00 \\
ETR & 77.5 & 5.0k & \underline{2.13} & 23.3 & 11.6k & 0.69 & 78.0 & 4.3k & 0.66 & 11.1 & 12.4k & \textbf{2.19} & 47.5 & 8.3k & \underline{1.15} \\
InfoDensity & 80.0 & 5.0k & \textbf{2.28} & 33.3 & 9.8k & \underline{2.31} & 80.2 & 3.4k & \textbf{0.91} & 9.6 & 11.8k & 1.54 & 50.8 & 7.5k & \textbf{1.49} \\

\midrule
\rowcolor{gray!20}
\multicolumn{16}{c}{\textit{DeepSeek-R1-Distill-Qwen-7B}} \\
\midrule
Original & 80.0 & 6.6k & 0.00 & 43.3 & 11.8k & 0.00 & 85.0 & 4.2k & 0.00 & 24.2 & 11.3k & 0.00 & 58.1 & 8.5k & 0.00
\\
DEER & 85.0 & 4.9k & 0.44 & 50.0 & 9.9k & 0.63 & 85.8 & 2.3k & 0.48 & 22.7 & 7.6k & 0.02 & 60.9 & 6.2k & 0.42 \\
NoThink & 77.5 & 3.8k & 0.27 & 56.7 & 8.3k & 1.22 & 81.4 & 1.7k & 0.38 & 22.2 & 2.0k & 0.41 & 59.5 & 4.0k & 0.60 \\
LCPO & 87.5 & 3.5k & 0.75 & 50.0 & 6.8k & 0.89 & 85.4 & 2.2k & 0.49 & 11.6 & 2.8k & -1.85 & 58.6 & 3.7k & 0.59 \\
O1-Pruner & 92.5 & 3.2k & \underline{0.98} & 46.7 & 7.8k & 0.57 & 89.0 & 2.1k & 0.64 & 39.4 & 6.2k & 2.34 & 66.9 & 4.8k & 0.89
\\
PEAR & 92.5 & 4.4k & 0.80 & 60.0 & 8.5k & 1.44 & 90.2 & 2.5k & 0.59 & 36.9 & 5.2k & 2.11 & 69.8 & 5.1k & 1.00 \\
ETR & 87.5 & 2.4k & 0.92 & 56.7 & 4.6k & \underline{1.54} & 90.6 & 1.5k & \textbf{0.84} & 37.3 & 2.5k & \underline{2.40} & 68.0 & 2.8k & \underline{1.18} \\
InfoDensity & 95.0 & 3.6k & \textbf{1.02} & 60.0 & 6.5k & \textbf{1.60} & 91.0 & 2.1k & \underline{0.70} & 42.9 & 5.6k &
\textbf{2.82} & 72.2 & 4.5k & \textbf{1.20} \\

\midrule
\rowcolor{gray!20}
\multicolumn{16}{c}{\textit{Qwen3-8B}} \\
\midrule
Original & 90.0 & 8.0k & 0.00 & 63.3 & 12.2k & 0.00 & 90.6 & 5.4k & 0.00 & 52.0 & 9.9k & 0.00 & 74.0 & 8.9k & 0.00 \\
DEER & 92.5 & 6.4k & 0.28 & 63.3 & 10.3k & 0.16 & 93.0 & 3.1k & 0.51 & 61.6 & 9.0k & 0.64 & 77.6 & 7.2k & 0.34 \\
NoThink & 67.5 & 3.4k & -0.68 & 40.0 & 7.0k & -1.41 & 86.4 & 1.5k & 0.49 & 26.8 & 4.4k & -1.87 & 55.2 & 4.1k & -0.73 \\
LCPO & 90.0 & 5.7k & 0.29 & 60.0 & 7.0k & 0.17 & 93.6 & 4.8k & 0.21 & 54.6 & 4.0k & 0.75 & 74.5 & 5.4k & 0.41 \\
O1-Pruner & 90.0 & 3.3k & \textbf{0.59} & 66.7 & 6.6k & 0.62 & 92.2 & 1.9k & \textbf{0.70} & 56.1 & 4.1k & \textbf{0.82} & 76.3 & 4.0k & \underline{0.64} \\
PEAR & 87.5 & 6.8k & 0.01 & 63.3 & 11.0k & 0.10 & 92.4 & 4.5k & 0.23 & 55.1 & 8.2k & 0.35 & 74.6 & 7.6k & 0.17 \\
ETR & 92.5 & 4.2k & \underline{0.56} & 73.3 & 8.6k & \underline{0.77} & 93.6 & 2.4k & 0.65 & 57.1 & 5.2k & 0.77 & 79.1 & 5.1k & 0.63 \\
InfoDensity & 90.0 & 3.7k & 0.54 & 73.3 & 6.9k & \textbf{0.91} & 93.0 & 2.1k & \underline{0.69} & 56.1 & 4.2k & \underline{0.81} & 78.1 & 4.2k & \textbf{0.69} \\
\bottomrule
\end{tabular}
}
\label{tab:baseline_t0_etr}
\end{table*}

\paragraph{Evaluation Metrics.}
We report three metrics. \textbf{Accuracy} (Acc) is pass@1 correctness on each benchmark. \textbf{Length} (Len) is the mean number of generated tokens per response. \textbf{Accuracy--Efficiency Score} (AES)~\citep{luo2025o1} jointly quantifies the accuracy--efficiency trade-off relative to the untrained base model:

\begin{equation}\label{eq:aes}
\mathrm{AES} = \begin{cases}
\alpha\,\Delta L + \beta\,|\Delta A|, & \text{if } \Delta A \geq 0, \\
\alpha\,\Delta L - \gamma\,|\Delta A|, & \text{if } \Delta A < 0,
\end{cases}
\end{equation}

where $A$ and $L$ denote accuracy (Acc) and response length (Len), and the deltas are computed relative to the untrained base model:
\[
\Delta L = \frac{L_\text{base} - L_\text{model}}{L_\text{base}},
\qquad
\Delta A = \frac{A_\text{model} - A_\text{base}}{A_\text{base}}.
\]
Following~\citet{luo2025o1}, we set $\alpha{=}1$, $\beta{=}3$, $\gamma{=}5$; the asymmetry $\gamma > \beta$ penalizes accuracy degradation more heavily than it rewards gains, discouraging methods that shorten reasoning at the cost of correctness. Higher AES indicates a better joint trade-off.

Further training and evaluation details are provided in Appendix~\ref{sec:appendix_experiments}.

\subsection{Experimental Results}
Table~\ref{tab:baseline_t0_etr} reports accuracy, response length, and AES across all methods and benchmarks.

\paragraph{Overall.}
InfoDensity achieves the highest overall AES on all three base models: $1.49$ on DSR-1.5B, $1.20$ on DSR-7B, and $0.69$ on Qwen3-8B. The largest margin appears on DSR-1.5B, where InfoDensity surpasses the strongest baseline (ETR, $1.15$) by $0.34$, a $30\%$ relative improvement. These gains are driven by simultaneous improvements in accuracy and reductions in length. On DSR-1.5B, mean accuracy rises from $36.5\%$ to $50.8\%$ ($+14.3$) while mean response length is compressed from $11.1$k to $7.5$k tokens ($-32\%$). The joint Acc/Len improvement holds on DSR-7B ($+14.1$/$-47\%$) and Qwen3-8B ($+4.1$/$-53\%$).

\paragraph{Per-dataset and baseline analysis.}
Across the 12 cells (4 benchmarks for 3 models), InfoDensity is the best in 6 and the second-best in 4, placing it in the top-2 for 10 of 12 cells. 
Its most consistent strength is on DSR-7B, where it wins 3 of 4 categories.
On DSR-1.5B, it wins AMC23 and MATH500, while PEAR edges it out on AIME24 ($2.68$ vs.\ $2.31$). 
On Qwen3-8B, where headroom for compression is smaller and all baselines tighten, InfoDensity still wins AIME24 and is second on MATH500 and GPQA-D, trailing O1-Pruner by only $0.01$. 

The baselines display distinct trade-offs. NoThink prioritizes length reduction at the cost of accuracy, losing $13.3$ and $23.3$ accuracy points on DSR-1.5B and Qwen3-8B AIME24 and yielding negative AES. LCPO and O1-Pruner achieve more modest length reductions. PEAR is competitive on the DeepSeek-R1-Distill family ($+1.00$ overall AES on both DSR-1.5B and DSR-7B) but less effective on Qwen3-8B ($+0.17$ overall AES), suggesting its effectiveness may depend on the underlying thinking-phase entropy structure. The concurrent ETR ranks second overall on DSR-1.5B and DSR-7B and third on Qwen3-8B, winning isolated cells (DSR-1.5B GPQA-D, DSR-7B MATH500).



\subsection{Effectiveness of InfoDensity} \label{sec:analysis_effectiveness}

\begin{table}[t]
\centering
\small
\setlength{\tabcolsep}{6pt}
\resizebox{\columnwidth}{!}{%
\begin{tabular}{l l c c}
\toprule
\makecell{Base \\ Model} & \makecell{After \\ \textsc{InfoDensity}} & Share & Mean length \\
\midrule
\multirowcell{2}[-0.7\normalbaselineskip]{Correct \\ ($n{=}387$)}
& Preserved   ($\checkmark$) & 92.2\%
& \makecell{$2471 \to 1622$ \\ ($-34\%$)} \\
\cmidrule(lr){2-4}
& Regression  ($\times$)     & 7.8\%
& \makecell{$4684\to 10323$ \\ ($+120\%$)} \\
\midrule
\multirowcell{2}[-0.7\normalbaselineskip]{Wrong \\ ($n{=}381$)}
& Rescued ($\checkmark$) & 27.6\%
& \makecell{$15228 \to 2700$ \\ ($-82\%$)} \\
\cmidrule(lr){2-4}
& Still wrong  ($\times$)        & 72.4\%
& \makecell{$14954 \to 12159$ \\ ($-19\%$)} \\
\bottomrule
\end{tabular}
}
\caption{Effect of \textsc{InfoDensity} on DSR-1.5B, pooled across four benchmarks. ``Mean length'' shows the mean response tokens change under $16384$-token generation cap.}
\label{tab:effectiveness}
\end{table}

We partition the test problems by whether the base model solves them correctly, and within each subgroup, record how InfoDensity changes the outcome in Table~\ref{tab:effectiveness}. Among the problems the base model already solves, InfoDensity preserves correctness on $92.2\%$ while cutting mean response length by $34\%$. By contrast, regressions are rare ($7.8\%$ of base-correct problems), and some are cases where InfoDensity itself exhausts the budget rather than producing short, confidently-wrong answers, which lengthens the mean response relative to the base. Among the problems the base model fails, InfoDensity rescues $27.6\%$, almost all of which are cases where the base model exhausts its $16384$-token budget without committing to an answer, while InfoDensity solves these problems in a mean of $2 700$ tokens. On problems where both methods fail, InfoDensity reduces mean length by only $19\%$ (compared to $34$--$82\%$ on cells where InfoDensity succeeds), indicating that compression is concentrated on problems InfoDensity can actually solve and that thinking budget is preserved on intrinsically difficult cases rather than being uniformly truncated. Representative paired generations for each outcome category are shown in Appendix~\ref{app:samples}.

\subsection{Length-Scaling Ablation}

\begin{figure}[tbh]
    \centering
    \includegraphics[width=0.9\linewidth]{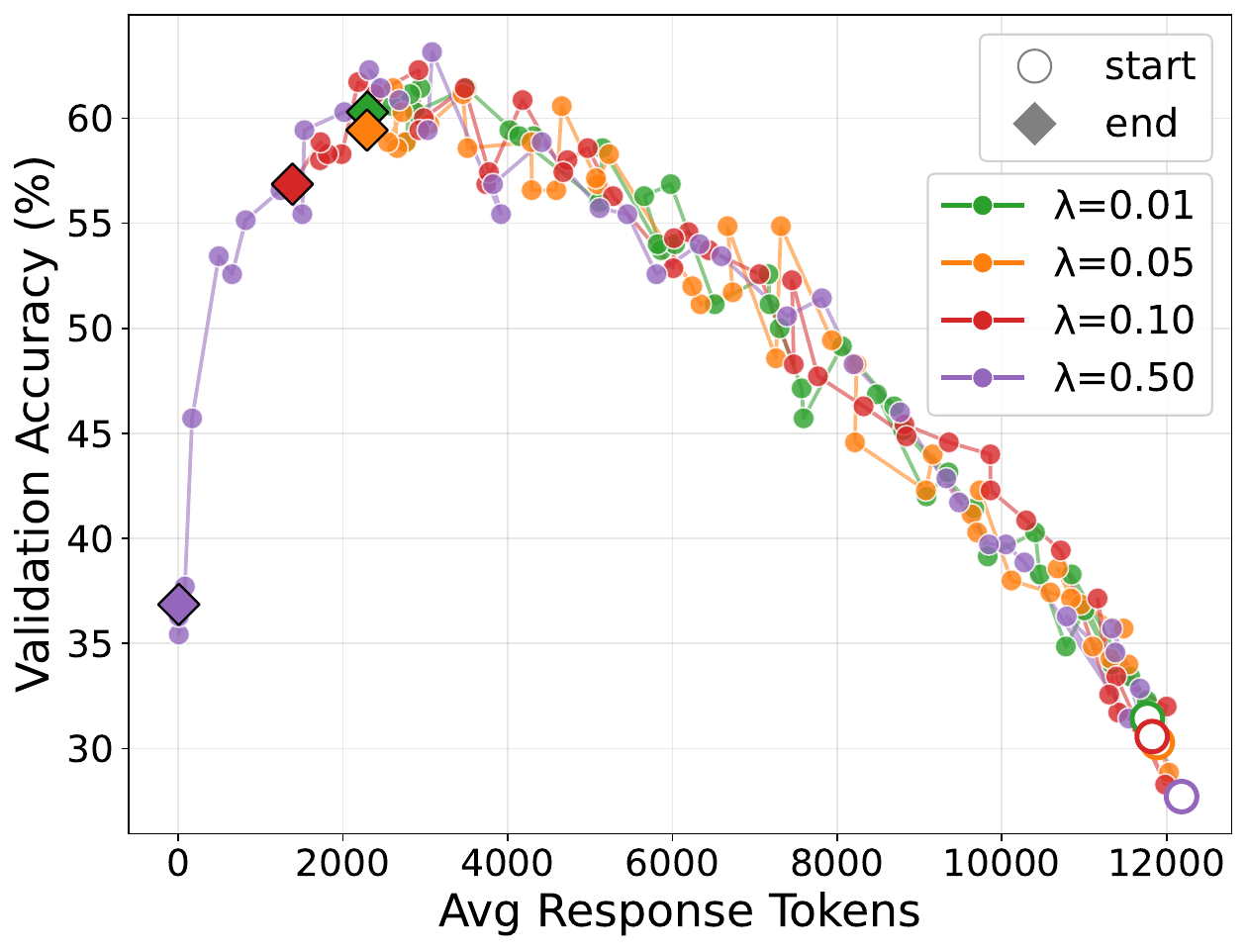}
    \caption{Length-scaling ablation on DSR-1.5B trained on DeepMath. Each trajectory traces the joint evolution of validation accuracy ($y$-axis) and average response length ($x$-axis) across training checkpoints, for four values of the length-scaling coefficient $\lambda$.}
    \label{fig:ablation}
\end{figure}

As $\lambda$ is the only hyperparameter specific to the InfoDensity reward, controlling the strength of length scaling, we ablate it across $\lambda \in \{0.01, 0.05, 0.1, 0.5\}$ on DSR-1.5B, with all four configurations trained for the same number of steps. Figure~\ref{fig:ablation} traces the joint evolution of validation accuracy and average response length across training checkpoints, starting from the untrained base model (circles) and ending at the final checkpoint (diamonds). The ideal direction is the upper-left corner: high accuracy with short responses.

We observe that the stability of InfoDensity training depends on $\lambda$. Moderate values ($\lambda \in \{0.01, 0.05, 0.1\}$) lead to stable and efficient endpoints, converging to validation accuracy of $57$--$60\%$ at response lengths of $1.4$--$2.5$k tokens. In contrast, the extreme setting $\lambda = 0.5$ causes the model to collapse: accuracy drops to around $37\%$, and response length shrinks dramatically, indicating that an excessive length penalty causes the model to prioritize extreme brevity over reasoning quality, which is a case of reward hacking. This pattern indicates that InfoDensity is robust to $\lambda$ across a broad range, with degradation appearing only at extreme length pressure.

\section{Conclusion}
We presented InfoDensity, a reward framework for RL-based training of LRMs that supervises reasoning quality through entropy trajectory signals. Through an empirical study on correct and incorrect reasoning traces, we identified two trajectory-level properties of high-quality reasoning: low uncertainty convergence and fast uncertainty descent. Motivated by this, InfoDensity uses a single suffix-max envelope of the entropy trajectory that jointly captures both properties, combined with a length scaling term that favors concise reasoning. Experiments show that InfoDensity outperforms state-of-the-art baselines on the joint accuracy--efficiency trade-off.

\section*{Limitations}
The ProcessBench traces analyzed in Section~\ref{sec:analysis} are not generated by the evaluated models that compute per-token entropies on them. We treat these traces as a proxy for characterizing each evaluated model's behavior on correct vs.\ incorrect reasoning, but the patterns observed on these external traces may not fully reflect those that would arise from each model's own generations.

Furthermore, as discussed in Section~\ref{sec:analysis} and Appendix~\ref{sec:appendix_qwen3_quirk}, compared with other models, the Qwen3 base models exhibit different entropy trajectory patterns on MATH and Omni-MATH. A definitive cause is beyond the scope of this work, but it suggests that the strength of the information-density signal may vary across model families and training distributions.


\bibliography{custom}

\appendix

\section{Implementation of Entropy Calculation}
\label{sec:appendix_entropy_impl}
\paragraph{Token entropy calculation.}
We obtain the per-token entropy $H(y_{t,k} \mid c_{t,k})$ defined in Section~\ref{sec:analysis_prelim} from vLLM~\cite{kwon2023efficient}. At every generated position vLLM returns the top-$K$ logprobs (we use $K{=}20$), and we compute the Shannon entropy
\[
H(y_{t,k} \mid c_{t,k}) \;=\; -\!\sum_{v \in \mathcal{V}_K} \tilde p_\theta(v \mid c_{t,k}) \log \tilde p_\theta(v \mid c_{t,k})
\]
over the top-$K$ vocabulary $\mathcal{V}_K \subset \mathcal{V}$, with the truncated probabilities $\tilde p_\theta(v \mid c_{t,k})$ renormalized to sum to one. Since the top-$K$ tail captures almost all of the probability mass for the high-confidence distributions characteristic of reasoning steps, the truncated estimate of $H(y_{t,k} \mid c_{t,k})$ closely approximates the full-softmax entropy while being upper-bounded by $\log K$, which later serves as a fixed normalization anchor in the reward in Section~\ref{sec:infodensity}.
\paragraph{From tokens to steps and chunks.}
For the ProcessBench analysis in Section~\ref{sec:analysis}, we delimit reasoning steps by the ``\texttt{\textbackslash n\textbackslash n}'' token, and compute $H_t$ and $\Delta H_t$ by token-level aggregation as in Section~\ref{sec:analysis_prelim}. For the RL training in Section~\ref{sec:infodensity}, we replace the step delimiter with fixed-size chunks of $C{=}50$ tokens because the ``\texttt{\textbackslash n\textbackslash n}'' boundary is unreliable when an exploring policy emits irregular formatting.

\section{Step-Level and Trajectory Analysis: Full 5-Model Panel}
\label{sec:appendix_step_level}
\begin{figure*}[tbh]
  \centering
  \includegraphics[width=0.98\linewidth]{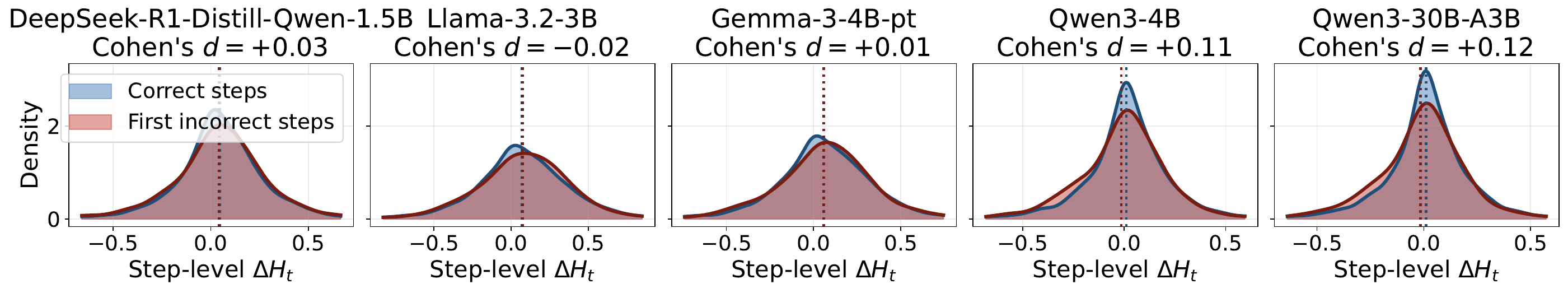}
  \caption{Per-step distributions of $\Delta H_t$ for correct (blue) and first-incorrect (red) steps in the ProcessBench subset, across the full 5-model panel: DeepSeek-R1-Distill-Qwen-1.5B, Llama-3.2-3B, Gemma-3-4B-pt, Qwen3-4B, and Qwen3-30B-A3B. Dotted vertical lines mark the per-class means; paired Cohen's $d$ values (printed in each panel) are uniformly small ($|d| \le 0.12$).}
  \label{fig:ig_dist_appendix}
\end{figure*}

\begin{figure}[tbh]
  \centering
  \includegraphics[width=0.85\linewidth]{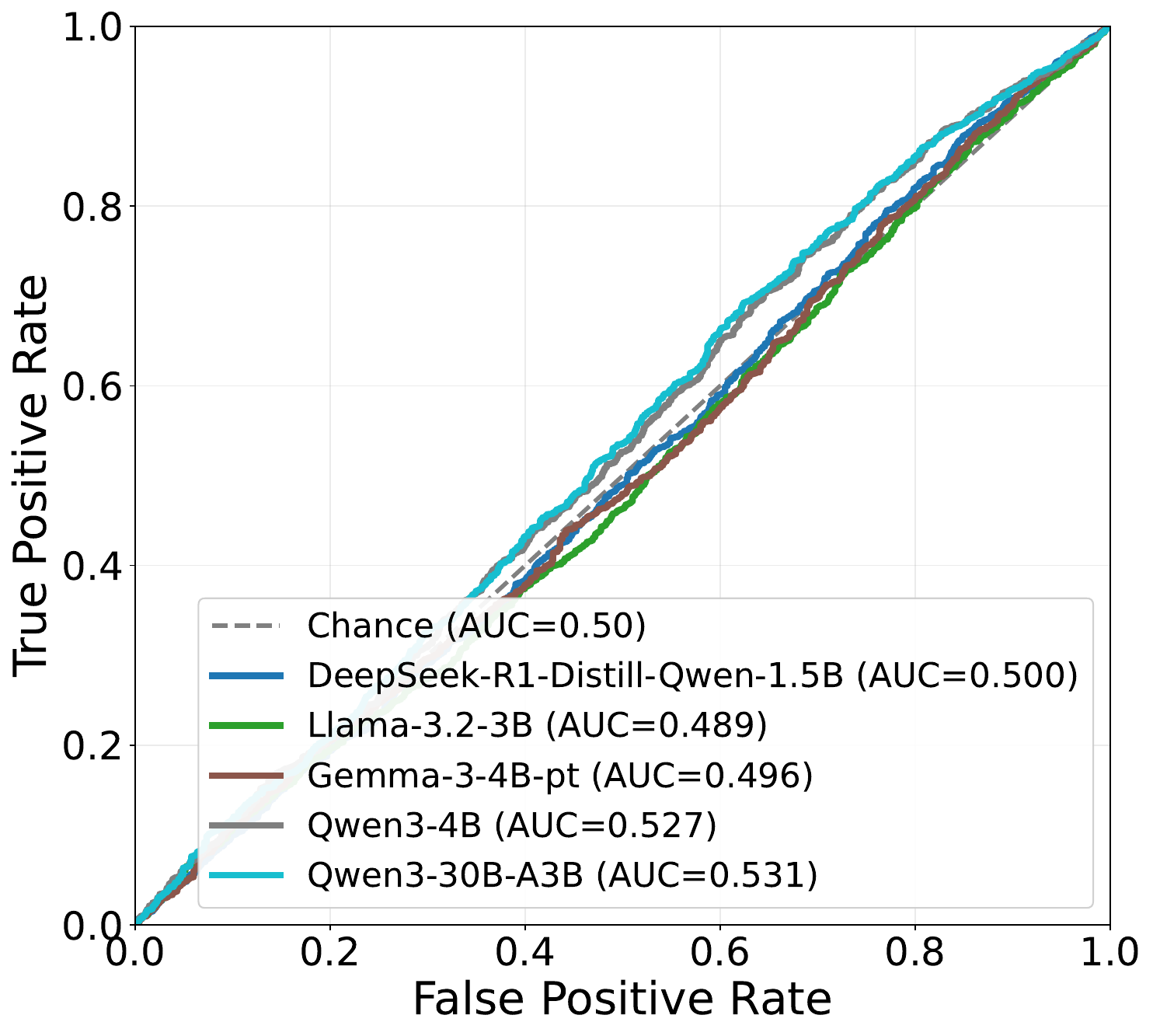}
  \caption{ROC curves for $\Delta H_t$ as a binary classifier of step correctness (positive class: correct steps; negative class: first-incorrect steps) in the ProcessBench subset across the full 5-model panel. All five curves cluster tightly around the chance diagonal (AUC $\in [0.489, 0.531]$), confirming that step-level $\Delta H_t$ is too noisy to serve as a robust error detector.}
  \label{fig:ig_roc}
\end{figure}

\begin{figure*}[tbh]
    \centering
    \includegraphics[width=0.95\linewidth]{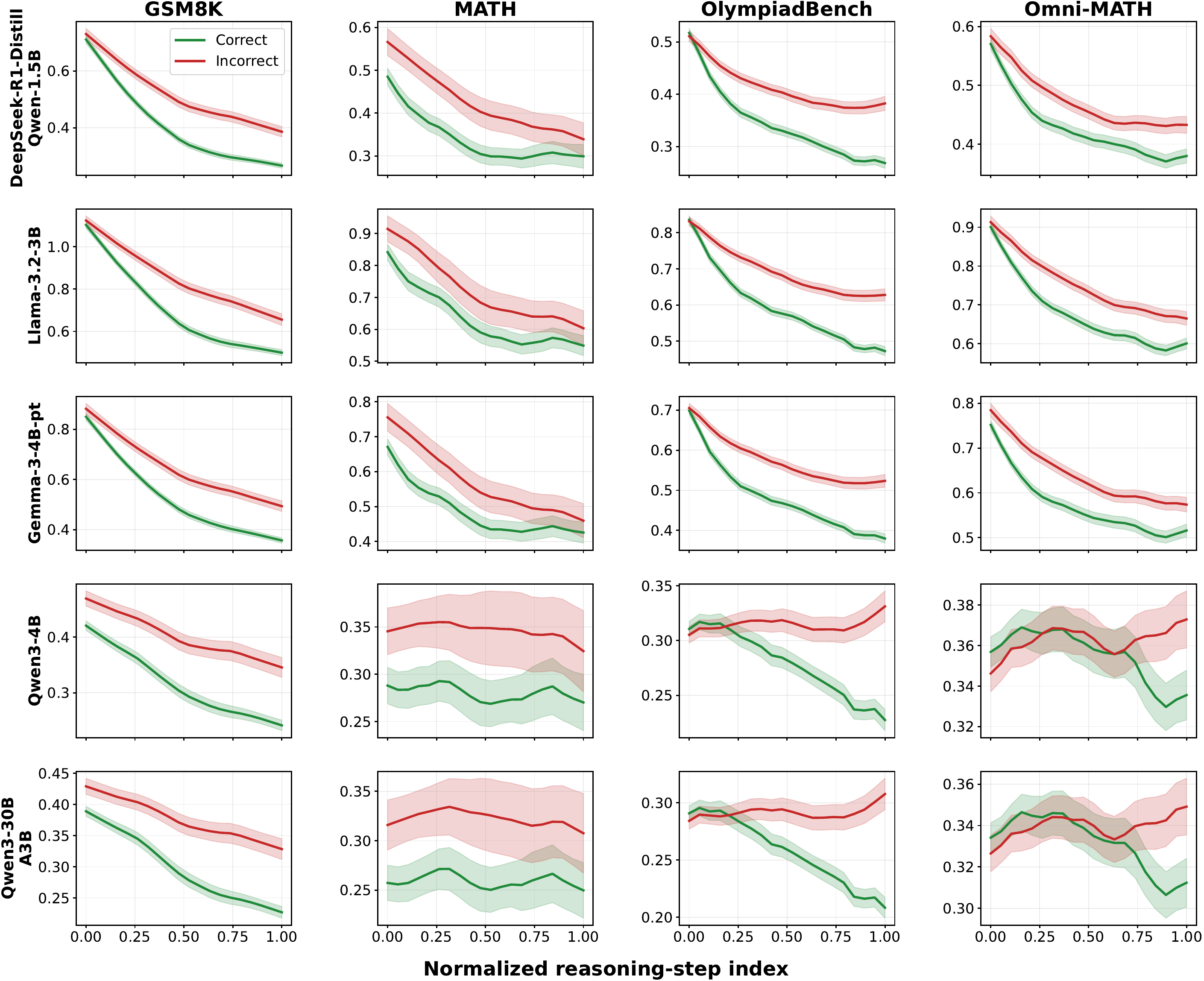}
    \caption{Step-level entropy trajectory of all five base models across four datasets: GSM8K, MATH, OlympiadBench, and Omni-MATH. Each curve shows the mean entropy over normalized reasoning steps, with shaded regions indicating $\pm 1$ standard error.}
    \label{fig:entropy_trajectory_5model}
\end{figure*}

This section shows the full 5-model panel for both the step-level $\Delta H_t$ analysis and the entropy trajectory analysis presented in Section~\ref{sec:ig_eval}. The KDE distributions of $\Delta H_t$ for correct vs.\ first-incorrect steps are in Figure~\ref{fig:ig_dist_appendix}, and the corresponding ROC curves in Figure~\ref{fig:ig_roc}. Both indicate that step-level $\Delta H_t$ is too noisy to discriminate correct from erroneous reasoning, with Cohen's $d \in [-0.02, +0.12]$ and AUC $\in [0.489, 0.531]$ across the five base models. The full five-model entropy trajectories are shown in Figure~\ref{fig:entropy_trajectory_5model}.

\section{Discussion on Qwen3 Entropy Trajectories}
\label{sec:appendix_qwen3_quirk}

In Section~\ref{sec:ig_eval} we observe that the two Qwen3 base models in our panel exhibit weaker entropy descent on MATH and Omni-MATH than the other three models, while showing the expected descent pattern on GSM8K and OlympiadBench. Non-Qwen models in the panel show consistent descent across all four datasets.
We examined several candidate explanations: problem-statement format, trace length, density of mathematical notation, and the distribution of generator models used to produce the ProcessBench traces. However, none cleanly accounts for the split. One remaining hypothesis is dataset-specific training-data overlap: the MATH benchmark and the Omni-MATH compilation draw heavily from older competition archives (AMC, AIME, and the like) widely present in pretraining corpora, while OlympiadBench's inclusion of newer physics-olympiad problems~\citep{he2024olympiadbench} and GSM8K's conversational word-problem format may invoke different generation regimes. Without access to Qwen3's pretraining data, we cannot conclusively verify this hypothesis.
This anomaly does not affect InfoDensity training. The suffix-max reward operates on within-rollout relative ranking across same-prompt completions rather than absolute entropy magnitudes, and the correct-vs-incorrect gap is preserved on the majority of cells (Section~\ref{sec:ig_eval}).

\section{Experimental Setup}
\label{sec:appendix_experiments}
\paragraph{Training and Evaluation.}
We train InfoDensity for one epoch (207 steps) on the 7000-problem DeepMath-103K subset described in Section~\ref{sec:infodensity}. We use LoRA with rank $r{=}32$ and scaling factor $\alpha{=}64$. During evaluation, following~\citet{xiong2026etr}, we decode with temperature $0$ (greedy) and a maximum generation length of $16384$ tokens across all benchmarks, and report pass@1 accuracy on AMC23, AIME24, MATH500, and GPQA-Diamond. Training and evaluation are conducted on NVIDIA H200 and A100 GPUs.
\paragraph{Baselines.}
We adopt the same base models, training data, and evaluation protocol as~\citet{xiong2026etr} throughout, so that our InfoDensity results are directly comparable to the numbers reported therein. Under this shared setup, we report the DSR-7B and Qwen3-8B baseline results directly from~\citet{xiong2026etr}, while for DSR-1.5B we train all baselines ourselves.

\section{Qualitative Samples}
\label{app:samples}

This appendix reproduces three representative paired generations from DSR-1.5B on MATH500, one from each of the behavioural categories identified in
the per-problem audit (Section~\ref{sec:analysis_effectiveness}):
a budget-exhaustion rescue, a clean compression of redundant
verification, and the failure mode of InfoDensity.
For each pair we report the question, ground truth, and key
excerpts from the Base and InfoDensity responses.
Both models use identical greedy decoding with the same prompt
template and a $16384$-token generation cap.

\subsection{Case 1: Budget exhaustion vs.\ commit (idx=345)}
\label{app:case_rescue}

\textbf{Question.} If $t(x) = 3 - g(x)$ and $g(x) = \sqrt{x}$,
what is $t(g(16))$?
\textbf{Ground truth.} $1$.

The problem is a two-step function composition that requires only
elementary arithmetic.  Base nevertheless fails to commit within
its budget: after solving the problem in the first
$\sim\!2000$ tokens, it enters a self-doubt loop that consumes
the remaining $14000$ tokens. InfoDensity solves the
same problem in $766$ tokens.

\begin{tcolorbox}[colback=gray!5, colframe=gray!50, breakable,
title={\textbf{Base} \,---\, tokens=$16384$, prediction=\detokenize{None}}]
\begin{Verbatim}[fontsize=\footnotesize, breaklines=true]
Okay, so I have this problem here: If t(x) = 3 - g(x) and g(x) = sqrt(x), then what is t(g(16))?  ...
[the model derives the correct intermediate value 1 within ~2,000 tokens, then enters the loop below]

... not correct because t(x) = 3 - g(x), so t(g(16)) = 3 - g(g(16)). Wait, no, that's not correct because t(x) = 3 - g(x), so t(g(16)) = 3 - g(g(16)). Wait, no, that's not correct because t(x) = 3 - g(x), so t(g(16)) = 3 - g(g(16)). Wait, no, that's not correct because ...
[the pattern repeats verbatim until the 16,384-token cap. No boxed answer is emitted]
\end{Verbatim}
\end{tcolorbox}

\begin{tcolorbox}[colback=orange!5, colframe=orange!50!black, breakable,
title={\textbf{\textsc{InfoDensity}} \,---\, tokens=$766$, prediction=$1$ $\checkmark$}]
\begin{Verbatim}[fontsize=\footnotesize, breaklines=true]
To find t(g(16)), we start by evaluating the inner function g(16). Since g(x) = sqrt(x), we have g(16) = sqrt(16) = 4.

Next, we evaluate t(4). Given t(x) = 3 - g(x), we substitute x = 4: t(4) = 3 - g(4).  We already know g(4) = sqrt(4) = 2.  Therefore, t(4) = 3 - 2 = 1.

Thus, the value of t(g(16)) is \boxed{1}.
\end{Verbatim}
\end{tcolorbox}

\subsection{Case 2: Redundant verification (idx=483)}
\label{app:case_compression}

\textbf{Question.} The area of triangle $ABC$ is
$a^2 - (b-c)^2$, where $a, b, c$ are the sides. Compute $\tan A$.
\textbf{Ground truth.} $\tfrac{8}{15}$.

Both methods reach the correct answer via the same opening
strategy: equate the given area with $\tfrac{1}{2}bc\sin A$,
use the Law of Cosines to express $\cos A$, and solve the
resulting quadratic for $\cos A=\tfrac{15}{17}$.  Base's response
contains a redundant second re-derivation of $\cos A$ via an
alternative algebraic path before committing, adding
$\sim\!6100$ thinking tokens while InfoDensity commits
after the first derivation.

\begin{tcolorbox}[colback=gray!5, colframe=gray!50, breakable,
title={\textbf{Base} \,---\, tokens=$8912$, prediction=$\tfrac{8}{15}$ $\checkmark$}]
\begin{Verbatim}[fontsize=\footnotesize, breaklines=true]
[opening derivation: equate area expressions, apply Law of Cosines, obtain quadratic in cos A]
...
[restart the derivation: expand (b-c)^2 again, re-apply Law of Cosines, re-derive the same quadratic 17*cos^2 - 32*cos + 15 = 0]
...
Solving: cos A = 1 or cos A = 15/17.  Since cos A = 1 is impossible, cos A = 15/17.  Then sin A = 8/17, so tan A = (8/17)/(15/17) = 8/15.

The value of tan A is \boxed{\frac{8}{15}}.
\end{Verbatim}
\end{tcolorbox}

\begin{tcolorbox}[colback=orange!5, colframe=orange!50!black, breakable,
title={\textbf{\textsc{InfoDensity}} \,---\, tokens=$2782$,
prediction=$\tfrac{8}{15}$ $\checkmark$}]
\begin{Verbatim}[fontsize=\footnotesize, breaklines=true]
[opening derivation: equate area expressions, apply Law of Cosines, obtain quadratic 17*cos^2 - 32*cos + 15 = 0]
...
Solutions: cos A = 1 and cos A = 15/17.  Since cos A = 1 is not possible, cos A = 15/17.  Then sin A = 8/17.

Thus, tan A = sin A / cos A = (8/17)/(15/17) = \boxed{\frac{8}{15}}.
\end{Verbatim}
\end{tcolorbox}

\subsection{Case 3: Arithmetic slip (idx=386, regression)}
\label{app:case_loss}

\textbf{Question.} How many integers are in the solution set of
$|x-2| \le 5.6$?  \textbf{Ground truth.} $11$.

Both methods identify the same compound inequality
$-5.6 \le x - 2 \le 5.6$ and isolate $x$ by adding $2$ to all
parts.  Base correctly computes $-5.6 + 2 = -3.6$ and lists
$11$ integers in $[-3.6, 7.6]$; InfoDensity mis-computes
$-5.6 + 2 = -2.6$ and lists $10$ integers in $[-2.6, 7.6]$.
No subsequent verification catches the slip.

\begin{tcolorbox}[colback=gray!5, colframe=gray!50, breakable,
title={\textbf{Base} \,---\, tokens=$1224$, prediction=$11$ $\checkmark$}]
\begin{Verbatim}[fontsize=\footnotesize, breaklines=true]
...  -5.6 <= x - 2 <= 5.6.  Adding 2 to all parts: -5.6 + 2 <= x <= 5.6 + 2,  so  -3.6 <= x <= 7.6.

The integers between -3.6 and 7.6 are -3, -2, -1, 0, 1, 2, 3, 4, 5, 6, 7  (11 integers).

\boxed{11}
\end{Verbatim}
\end{tcolorbox}

\begin{tcolorbox}[colback=orange!5, colframe=orange!50!black, breakable,
title={\textbf{\textsc{InfoDensity}} \,---\, tokens=$878$, prediction=$10$ $\times$}]
\begin{Verbatim}[fontsize=\footnotesize, breaklines=true]
...  -5.6 <= x - 2 <= 5.6.  Adding 2 to all parts: -5.6 + 2 <= x <= 5.6 + 2,  so  -2.6 <= x <= 7.6.   <-- arithmetic slip

The integers in this range are -2, -1, 0, 1, 2, 3, 4, 5, 6, 7 (10 integers).

\boxed{10}
\end{Verbatim}
\end{tcolorbox}

This miscount illustrates a common failure mode of InfoDensity on MATH500: a single arithmetic slip, most commonly a sign or counting error, rather than a breakdown in higher-level reasoning.

\end{document}